\documentclass{article} 

\usepackage{graphicx}
\usepackage{subcaption}
\usepackage{amsmath}
\usepackage{amssymb}
\usepackage{algorithm,algpseudocode}
\usepackage{nips15submit_e,times}
\usepackage{hyperref}
\usepackage{url}
\usepackage{multirow}

\title{Seq-NMS for Video Object Detection}

\author{
Wei Han$^{1}$\thanks{Authors contributed equally to this work}, Pooya Khorrami$^{1}$\footnotemark[1], Tom Le Paine$^{1}$\footnotemark[1], Prajit Ramachandran$^{1}$,\\
\textbf{Mohammad Babaeizadeh$^{1}$, Honghui Shi$^{1}$, Jianan Li$^{2}$} \\
\textbf{Shuicheng Yan$^{2}$, Thomas S. Huang$^{1}$}\\
\\
$^{1}$University of Illinois at Urbana-Champaign \\
\texttt{\{weihan3, pkhorra2, paine1, prmchnd2} \\
\texttt{mb2, hshi10, t-huang1\}@illinois.edu} \\
$^{2}$National University of Singapore \\
\texttt{\{elev373, eleyans\}@nus.edu.sg}
}

\nipsfinalcopy 

\begin{document}
\maketitle
                                                                                                                                                                                                                                                                                                                                                                                                                                                                                                                                                                                                                                                                                                           
\begin{abstract}
Video object detection is challenging because objects that are easily detected in one frame may be difficult to detect in another frame within the same clip. Recently, there have been major advances for doing object detection in a single image. These methods typically contain three phases: (i) object proposal generation (ii) object classification and (iii) post-processing. We propose a modification of the post-processing phase that uses high-scoring object detections from nearby frames to boost scores of weaker detections within the same clip. We show that our method obtains superior results to state-of-the-art single image object detection techniques. Our method placed $3^{rd}$ in the video object detection (VID) task of the ImageNet Large Scale Visual Recognition Challenge 2015 (ILSVRC2015). 
\end{abstract}

\section{Introduction}
Single image object detection has experienced large performance gains in the last few years. Video object detection, on the other hand, still remains an open problem. This is mainly because objects that are easily detected in one frame may be difficult to detect in another frame within the same video clip. There are many reasons that may cause this difficulty. Some examples include: (i) drastic scale changes (ii) occlusion and (iii) motion blur. In this work we propose a simple extension of single image object detection to help overcome these difficulties. 

Current state-of-the-art single image object detection systems can be broken up into three distinct phases: (i) region proposal generation (ii) object classification and (iii) post-processing. During the region proposal generation phase, a set of candidate regions are generated based on how likely they are to contain an object. Previous region proposal methods were based on low-level image features \cite{selectivesearch, edgeboxes} while the current state-of-the-art, Faster R-CNN, \cite{ren15fasterrcnn} learns to generate proposals using a neural network. The candidate regions are then assigned a class score in the object classification phase, and redundant detections are subsequently filtered in the post-processing phase.




While effective, single image methods are na\"{i}ve because they completely ignore the temporal dimension. In this work, we incorporate temporal information during the post-processing phase in order to refine the detections within each individual frame. Given a video sequence of region proposals and their corresponding class scores, our method associates bounding boxes in adjacent frames using a simple overlap criterion. It then selects boxes to maximize a sequence score. Those boxes are then used suppress overlapping boxes in their respective frames and are subsequently rescored in order to boost weaker detections.

The main contributions of our work are as follows:
\begin{enumerate}
\item We present Seq-NMS, a method to improve object detection pipelines for video data. Specifically, we modify the post-processing phase to use high-scoring object detections from nearby frames in order to boost scores of weaker detections within the same clip.

\item We evaluate Seq-NMS on the ImageNet VID dataset and show that it outperforms state-of-the-art single image-based methods. We show that our method is helpful in cases where single frames contain objects that are at extreme scales, occluded, or blurred. We present specific instances where our Seq-NMS improves performance.

\item Our method placed $3^{rd}$ in the video object detection (VID) task of the ImageNet Large Scale Visual Recognition Challenge 2015 (ILSVRC2015). 
\end{enumerate}

\section{Our Approach}
\begin{figure}[t!]
  \centering
  \includegraphics[width=1.0\textwidth]{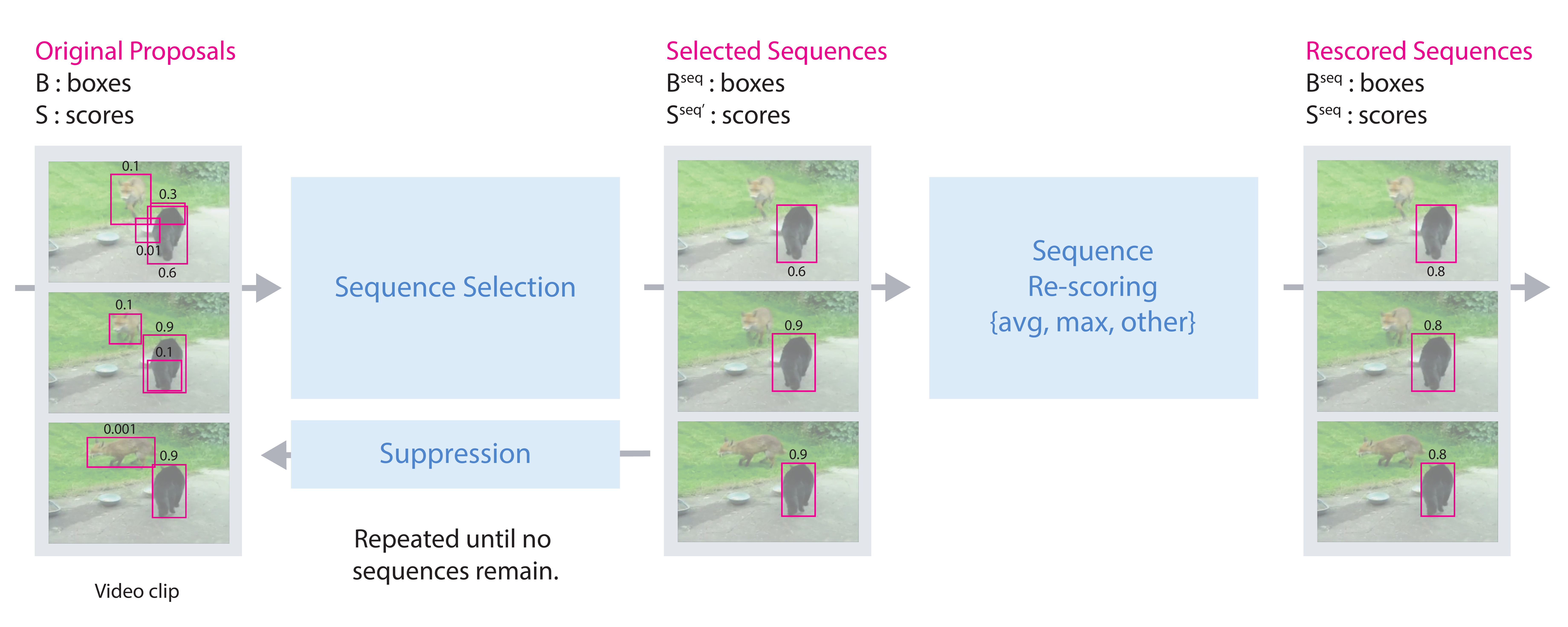}
  \caption{Illustration of Seq-NMS. Seq-NMS takes as input all object proposals boxes \textbf{B} and scores \textbf{S} for an entire video clip \textbf{V} (in contrast to NMS which takes proposals from a single image). It is applied iteratively. At each iteration it performs three steps: 1) \textbf{Sequence Selection}, which selects the sequence of boxes with the highest sequence score, $\textbf{B}^{seq}$. 2) \textbf{Sequence Re-scoring}, which takes all scores in the sequence $\textbf{S}^{seq'}$ and applies a function to them to get a new score for each frame in the sequence $\textbf{S}^{seq}$. 3) \textbf{Suppression}, which for each box in $\textbf{B}^{seq}$, suppresses any boxes in the same frame that have sufficient overlap.}
  \label{fig:flow_chart}
\end{figure}

\begin{figure}[t!]
  \centering
  \includegraphics[width=1.0\textwidth]{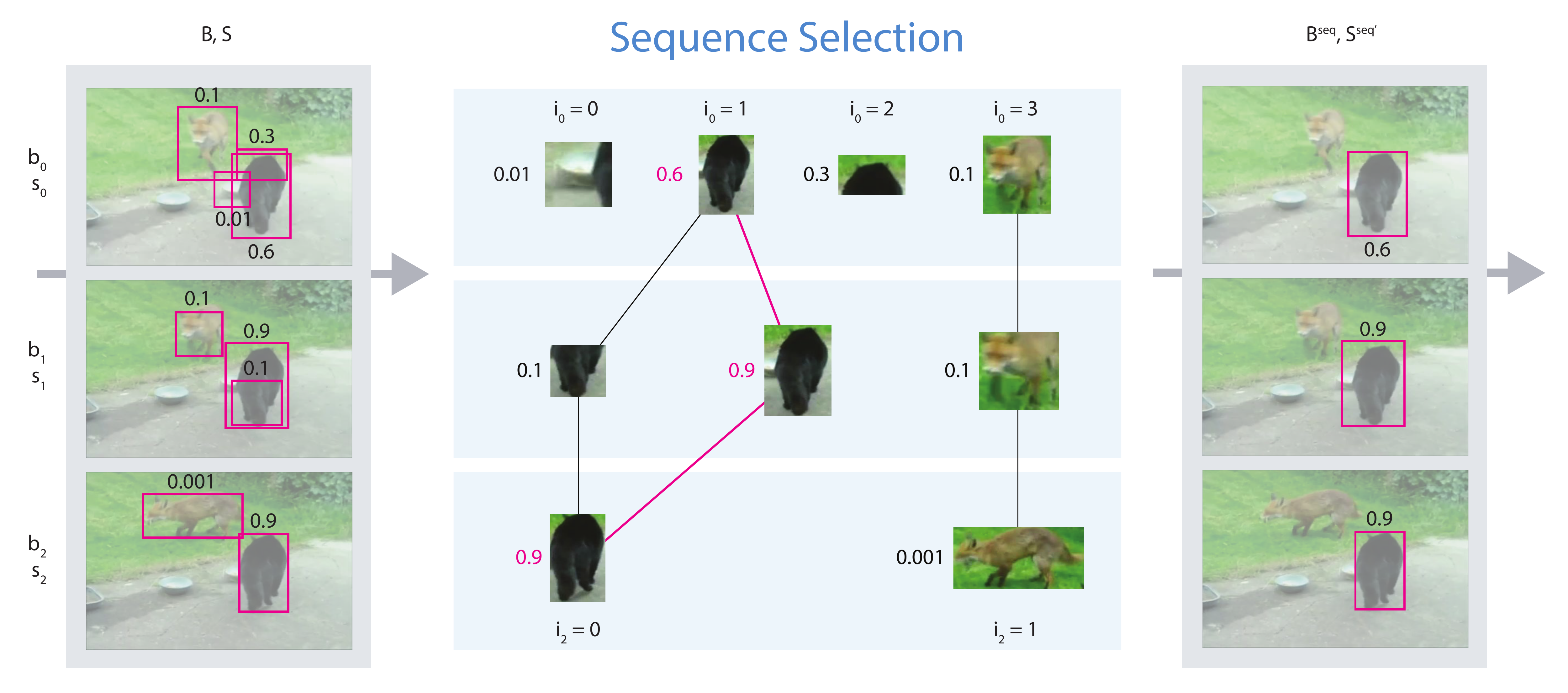}
  \caption{
Illustration of Sequence Selection. We construct a graph where boxes in adjacent frames are linked iff their IoU $>$ 0.5. A sequence is a set of boxes that are linked in the video. We then select the sequence with the highest sequence score shown in Equation \ref{eq:seq_select_cost}.
This produces $\textbf{B}^{seq}$ and $\textbf{S}^{seq'}$ which is a set of at most one box per frame, and the associated scores. After Sequence Selection, for each box in $\textbf{B}^{seq}$, we suppress any boxes in the same frame that have IoU $>$ 0.3.}
  \label{fig:seq_nms_example}
\end{figure}



\subsection{Seq-NMS}

Most object detection methods (Faster R-CNN included) are designed for performing object detection on a single independent frame. However, since we are concerned with object detection in videos, it would be a waste of salient information to ignore the temporal component entirely. One problem we noticed with Faster R-CNN on the validation set was that non-maximum suppression (NMS) frequently chose the wrong bounding box after object classification. It would choose boxes that were overly large, resulting in a smaller intersection-over-union (IoU) with the ground truth box because the union of areas was large. The large boxes often had very high object scores, possibly because more information is available to be extracted during RoI pooling. In order to combat this problem, we attempted to use temporal information to re-rank boxes. We assume that neighboring frames should have similar objects, and their bounding boxes should be similar in position and size, i.e. temporal consistency. 

To make use of this temporal consistency, we propose a heuristic method for re-ranking bounding boxes in video sequences called Seq-NMS. Seq-NMS has three steps: Step 1) \textbf{Sequence Selection}, Step 2) \textbf{Sequence Re-scoring}, Step 3) \textbf{Suppression}. We repeat these three steps until a no sequences are left. Figure \ref{fig:flow_chart} illustrates this process.

Seq-NMS is performed on a single video clip \textbf{V} which is comprised of a set of $T$ frames, $\{ v_0, \ldots, v_T \}$. For each frame $t$, we have a set of region proposal boxes $b_t$ and scores $s_t$ both of size $n_t$, which varies for each frame. The set of proposals for an entire clip is denoted by $\textbf{B} = \{b_0, \ldots, b_T \}$. Likewise, the set of scores for the entire clip is denoted by $\textbf{S} = \{s_0, \ldots, s_T \}$.

Given a set of region bounding boxes $\textbf{B}$, and their detection scores $\textbf{S}$ as input, sequence selection chooses a subset of boxes $\textbf{B}^{seq}$ and their associated scores $\textbf{S}^{seq'}$. The re-scoring function takes $\textbf{S}^{seq'}$ and produces a new set of scores $\textbf{S}^{seq}$.

\textbf{Sequence Selection.} For each pair of neighboring frames in \textbf{V}, a bounding box in the first frame can be linked with a bounding box in the second frame iff their IoU is above some threshold. We find potential linkages in each pair of neighboring frames across the entire clip. Then, we attempt to find the maximum score sequence across the entire clip. That is, we attempt to find the sequence of boxes that maximize the sum of object scores subject to the constraint that all adjacent boxes must be linked.
\begin{eqnarray}
\label{eq:seq_select_cost}
i' &=& \underset{i_{t_s},\ldots,i_{t_e}}{\operatorname{argmax}} \sum_{t=t_s}^{t_e} s_t[i_t] \nonumber \\
&s.t.& 0\leq t_s \leq t_e < T \\
&s.t.& IoU(b_t[i_t], \: b_{t+1}[i_{t+1}]) > 0.5, \: \forall t \in [t_s, t_e)  \nonumber
\end{eqnarray}

This can be found efficiently using a simple dynamic programming algorithm that maintains the maximum score sequence so far at each box. The optimization returns a set of indices $i'$ that are used to extract a sequence of boxes $\textbf{B}^{seq} = \{b_{t_s}[i_{t_s}], \ldots, b_{t_e}[i_{t_e}]\}$ and their scores $\textbf{S}^{seq'} = \{s_{t_s}[i_{t_s}], \ldots, s_{t_e}[i_{t_e}]\}$. Figure \ref{fig:seq_nms_example} gives a visual example of the sequence selection phase. 



\textbf{Sequence Re-scoring.} After the sequence is selected, the scores within it are improved. We apply a function $F$ to the sequence scores to produce $\textbf{S}^{seq} = F(\textbf{S}^{seq'})$. We try two different re-scoring functions: the average and the max. 

\textbf{Suppression.} The boxes in the sequence are then removed from the set of boxes we link over. Furthermore, we apply suppression within frames such that if a bounding box in frame $t, t \in [t_s, t_e]$, has an IoU with $b_t$ over some threshold, it is also removed from the set of candidate boxes. 

\section{The Dataset}
For the 2015 iteration, the ImageNet competition contained a new taster competition for object detection from video called the ImageNet VID competition. Similar to the ImageNet object detection task (DET), the task is to classify and locate objects in every image. However, instead of containing a collection of independent images, the VID dataset groups several frames from the same video into video clips or "snippets". All visible objects in every frame are annotated with a class label and bounding box. The VID dataset contains 30 object categories which are a subset of the 200 categories provided in the DET dataset. The dataset contains three sets of non-overlapping videos and labels: train, validation, and test. The training, validation and test sets in the initial release of the VID dataset contain 1,952, 281 and 458 snippets respectively. Meanwhile, the final release roughly doubled the number snippets in each set to 3,862, 555, and 937. The number of snippets and number of images in each set of the ImageNet VID dataset can be found in Table \ref{table:vid_dataset_numbers}.

\begin{table}[t!]
\centering
\caption{Number of Samples in Imagenet VID Dataset}
\label{table:vid_dataset_numbers}
\begin{tabular}{cc|c|c|c|}
\cline{3-5}
& & Train & Validation & Test \\ 
\hline
\multicolumn{1}{ |c  }{\multirow{2}{*}{Initial} } &
\multicolumn{1}{ |c| }{Snippets} & 1,952 & 281 & 458 \\ 
\multicolumn{1}{ |c  }{}  &
\multicolumn{1}{ |c| }{Images} & 405,014 & 64,698 & 127,618 \\ 
\hline 
\multicolumn{1}{ |c  }{\multirow{2}{*}{Full} } &
\multicolumn{1}{ |c| }{Snippets} & 3,862 & 555 & 937 \\ 
\multicolumn{1}{ |c  }{}  &                     
\multicolumn{1}{ |c| }{Images} & 1,122,397 & 176,126 & 315,176 \\ \cline{1-5}
\end{tabular}
\end{table}

\section{Results}
\subsection{Training Details for RPN and Classifier}

In Faster R-CNN, the RPN and the classification network share convolutional layers and are trained together in an alternating fashion. First, we trained a Zeiler Fergus (ZF) style \cite{viz_und} RPN using stochastic gradient descent and the image sampling strategy described in \cite{fastrcnn}. We accomplished this by first training the RPN on the initial VID training dataset for 400,000 iterations. We then trained a ZF style Fast R-CNN on the initial VID training set for 200,000 iterations. Finally, we refined the RPN by fixing the convolutional layers to be those of the trained detector and trained for 400,000 steps. We found that our trained RPN was able to obtain proposals that overlapped with the ground truth boxes in the initial VID validation set with recall over 90\%. 

For our classifier, we considered both a Zeiler Fergus style network (ZF net) and VGG16 network (VGG net) \cite{simonyan2014very}. The ZF network was trained on the initial VID training set and the VGG16 net was pre-trained on the training and validation sets of the 2015 ImageNet DET challenge. The DET dataset contained 200 object categories and the train and validation sets contained 456,567 and 55,502 images, respectively. We then replaced the 200 unit softmax layer with a 30 unit one and trained it on the initial VID training set (405K images) while keeping all of the other layers fixed. It should be noted that we never used the full training set (1.1M images) in any of our experiments. Our models were trained using a heavily modified version of the open source Faster R-CNN Caffe code released by the authors\footnote{https://github.com/rbgirshick/py-faster-rcnn}.


\subsection{Quantitative Results}
We validated our method by conducting experiments on the initial and full validation set as well as the full test set of the ImageNet VID dataset. During the post-processing phase, we considered four different techniques: (i) single image NMS (ii) Seq-NMS (avg) (iii) Seq-NMS (max) (iv) Seq-NMS (best). Seq-NMS (avg) and Seq-NMS (max) rescored the sequences selected by Seq-NMS using the average or max detection scores respectively, while Seq-NMS (best) chose the best performing of the three aforementioned techniques on each class and averaged the results.

\begin{figure}[t!]
  \centering
  \includegraphics[width=1.0\textwidth]{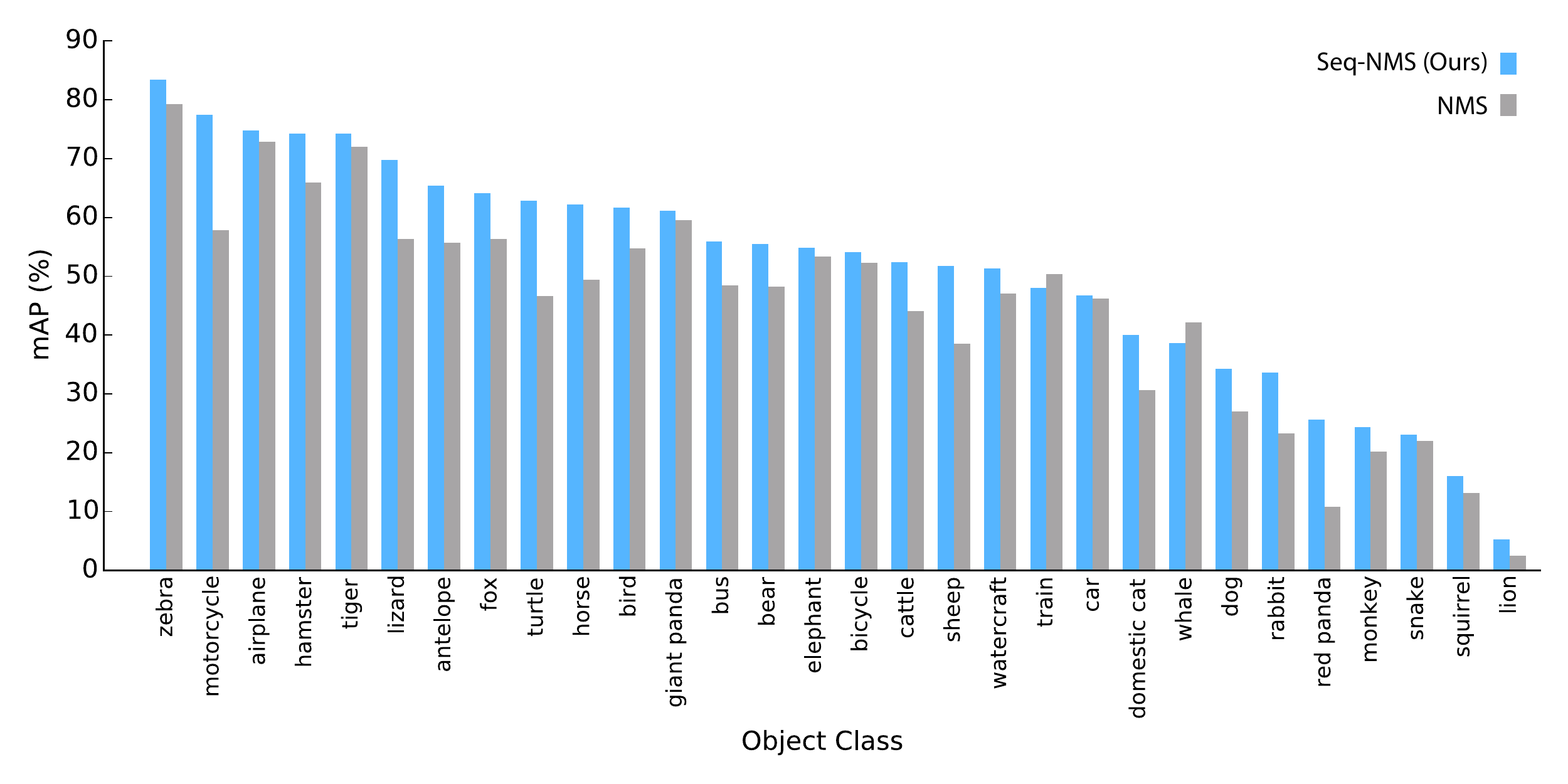}
  \caption{Performance (mAP) of our Seq-NMS and NMS. Performance is measured on the full ImageNet validation set. We use average rescoring for Seq-NMS. The classes are sorted in descending order by Seq-NMS performance.}
  \label{fig:performance_bar_chart}
\end{figure}

\begin{figure}[t!]
  \centering
  \includegraphics[width=1.0\textwidth]{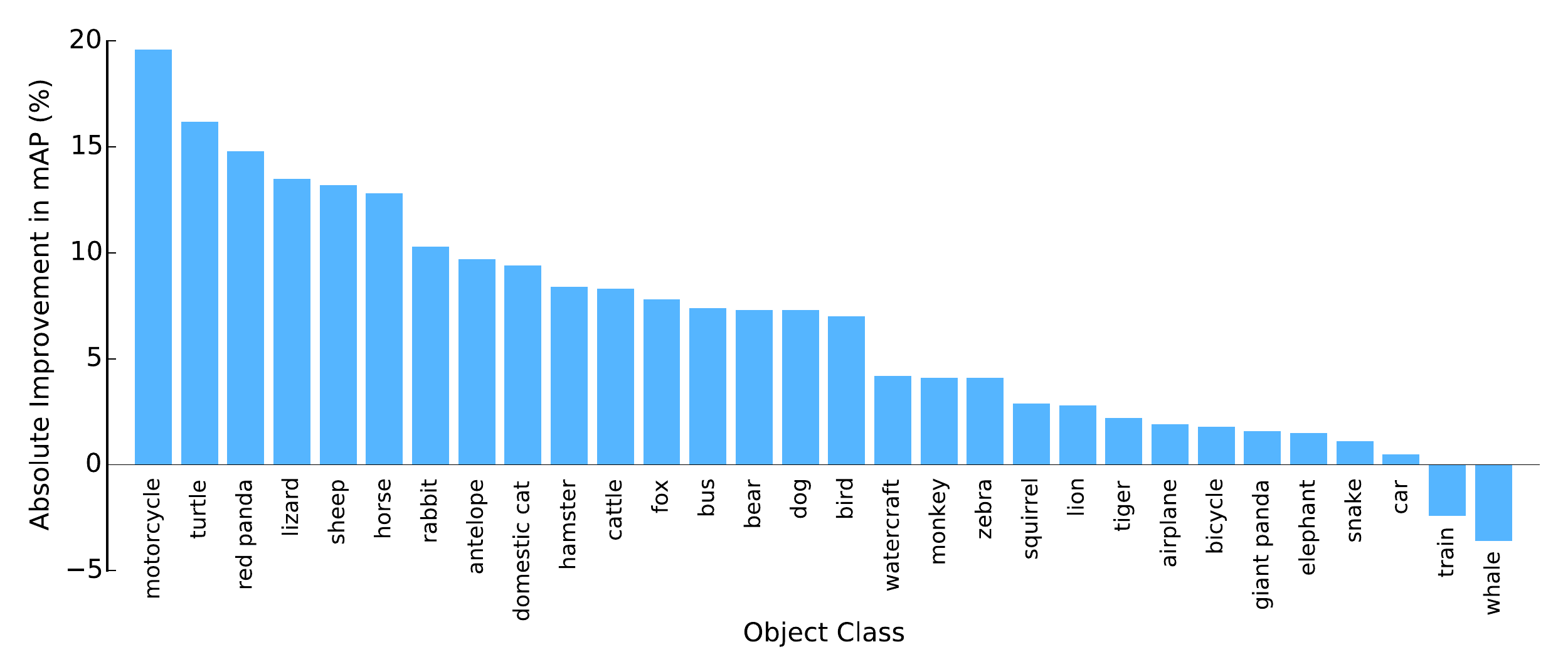}
  \caption{Absolute improvement in mAP (\%) using Seq-NMS. The improvement is relative to single image NMS. Note that 7 classes have higher than 10\% improvement, and only two classes show decreased performance (train and whale).}
  \label{fig:class_improvement_bar_chart}
\end{figure}

Table \ref{table:method_comp_init_val} shows our results on the initial and full validation set. We found that using VGG net gave a substantial improvement over using the architecture described by Zeiler and Fergus. Sequence re-scoring with Seq-NMS gave further improvements. On the initial validation set, Seq-NMS (avg) achieved a mAP score of $51.5\%$. This result can be further improved to $53.6\%$ when combining all three NMS techniques. Meanwhile on the full validation set, Seq-NMS (avg) got a mAP score of $51.4\%$. When combining all three NMS methods (Seq-NMS (best)) on the full val set, we achieve a mAP score of $52.2\%$.  In Figure \ref{fig:performance_bar_chart}, we give a full breakdown of Seq-NMS' (avg) performance across all 30 classes and compare it with the single image NMS technique. Figure \ref{fig:class_improvement_bar_chart} shows which classes experienced the largest gains in performance when switching from single image NMS to Seq-NMS (avg). The 5 classes that experienced the highest gains in performance were: (i) motorcycle (ii) turtle (iii) red panda (iv) lizard and (v) sheep.

On the test set, we ranked $3^\text{rd}$ in terms of overall mean average precision (mAP). The results of VGG net models are shown in Table \ref{table:mAP_test_set}. Once again, we see that Seq-NMS and Rescoring showed significant improvements over traditional frame-wise NMS post-processing. Our best submission achieved a mAP of $48.7\%$\footnote{http://image-net.org/challenges/LSVRC/2015/results}.

We also report the mAP score of the challenge's top performing method \cite{kang2016t} on both the validation and test sets in Tables \ref{table:method_comp_init_val} and \ref{table:mAP_test_set}. In \cite{kang2016t}, the authors present a suite of techniques including (i) a strong still-image detector (ii) bounding box suppression and propagation (iii) trajectory / tubelet re-scoring and (iv) model combination. The still-image detector's performance achieves a mAP of 67.7\%. When directly comparing the amount of improvement obtained just from temporal information (ii) and (iii), our method is superior (7.3\% vs. 6.7\%).

\subsection{Qualitative Results}
\begin{figure}[t!]
  \centering
  \includegraphics[width=1.0\textwidth]{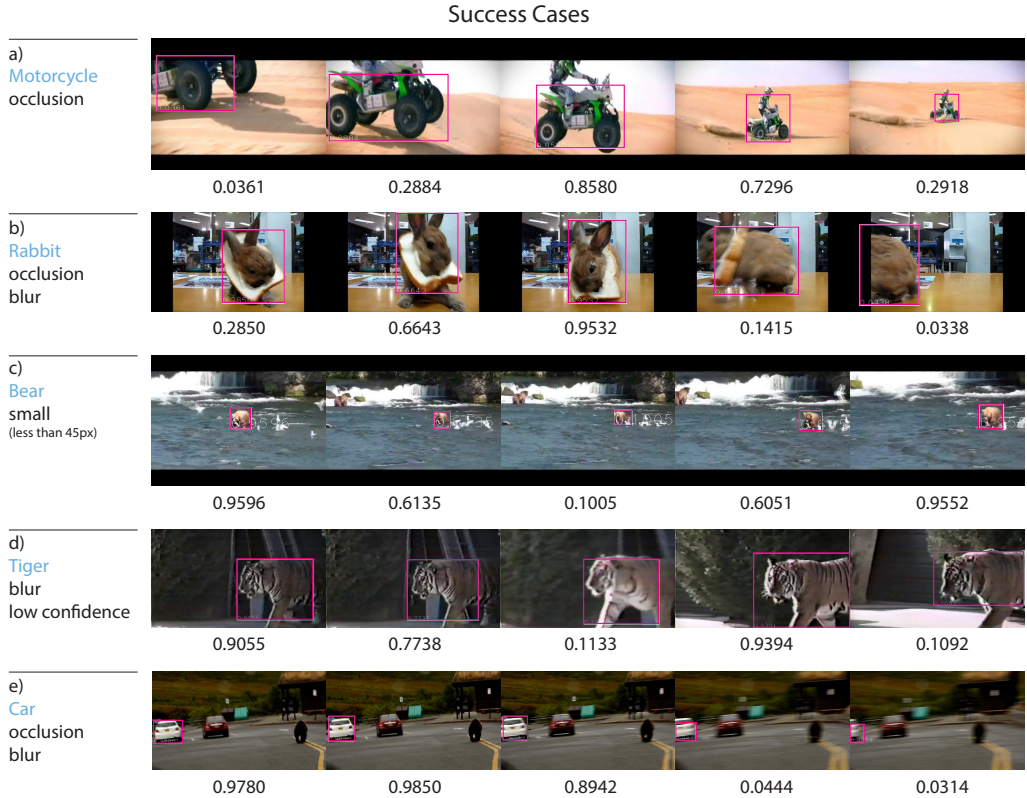}
\caption{Example video clips where Seq-NMS improves performance. 
  The boxes represent a sequence selected by Seq-NMS. Clips are subsampled to provide examples of high and low scoring boxes. In clips \textbf{a}, \textbf{b}, and \textbf{e}, the object becomes more and more occluded as it exits the frame, leading to lower scores. Meanwhile, in clips \textbf{c} and \textbf{d}, the object of interest has a low classifier score because it is either very small or blurred, respectively. In all of these cases, Seq-NMS' rescoring significantly boosts the weaker detections by using the strong detections from adjacent frames.}
  \label{fig:success_cases}
\end{figure}

In Figure \ref{fig:success_cases}, we present clips from the ImageNet VID dataset where Seq-NMS improved performance. The boxes represent a sequence selected by Seq-NMS. Clips were subsampled to provide examples of high and low scoring boxes. In each of these clips, the object of interest is subjected to one or more perturbations commonly seen in video data such as  occlusion (clips a, b, and e), drastic scaling (clip c), and blur (clip d). These perturbations naturally cause the classifier to score proposals with much lower confidence. However, since the Seq-NMS has associated these lower confidence detections with previous higher confidence detections of the same object, rescoring the lower confidence detections with the average improves precision. 

\begin{figure}[t!]
  \centering
  \includegraphics[width=1.0\textwidth]{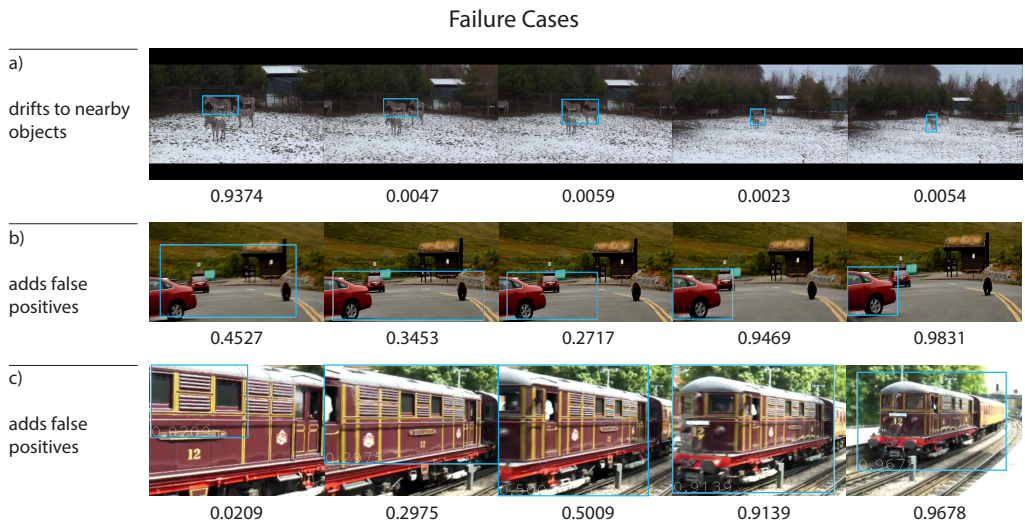}
  \caption{Video clips in the ImageNet VID dataset where Seq-NMS does not improve performance. In clip \textbf{a}, Seq-NMS has difficulty when there are several objects with similar appearance close together in the video (clip \textbf{a}). This will cause the detector to drift from one object to another which leads to missed detections and incorrect score assignment. Seq-NMS also accumulates spurious detections which leads to many false positives (clips \textbf{b} and \textbf{c}). This occurs because Seq-NMS' objective function does not penalize against adding more detections.}
  \label{fig:failure_cases}
\end{figure}

We also present instances where Seq-NMS does not appear to improve performance in Figure \ref{fig:failure_cases}. One case where Seq-NMS may not help is when there are several objects with similar appearance close together in the video (clip a). This will cause the detector to drift from one object to another which leads to missed detections and incorrect score assignment. Another case is when Seq-NMS accumulates spurious detections which leads to many more false positives (clips b and c). This occurs because Seq-NMS' objective function, the sum of a sequence's confidence scores, does not penalize against adding more detections.

\begin{table}[t!]
\centering
\caption{Method comparison on initial and full ImageNet VID validation set}
\label{table:method_comp_init_val}
\begin{tabular}{|l|c|c|}
\hline
\textbf{Method} & \textbf{mAP(\%) - (Initial Val)} & \textbf{mAP(\%) - (Full Val)} \\ \hline
ZF net + NMS & 32.2 & - \\ 
ZF net + Seq-NMS (max) & 36.3 & - \\ 
ZF net + Seq-NMS (avg) & 38.3 & - \\ 
ZF net + Seq-NMS (best) & 40.2 & - \\ \hline
VGG net + NMS & 44.4 & 44.9 \\ 
VGG net + Seq-NMS (max) & 50.1 & 50.5 \\ 
VGG net + Seq-NMS (avg) & 51.5 & 51.4 \\ 
VGG net + Seq-NMS (best) & 53.6 & 52.2 \\ \hline
\textbf{CUVideo - T-CNN \cite{kang2016t}} & \textbf{-} & \textbf{73.8} \\ \hline
\end{tabular}
\end{table}

\begin{table}[t!]
\centering
\caption{Method comparison on full ImageNet VID test set}
\label{table:mAP_test_set}
\begin{tabular}{|l|c|}
\hline
\textbf{Method} & \textbf{mAP (\%)} \\ \hline
VGG net + NMS & 43.4 \\ 
VGG net + Seq-NMS (max) & 47.5 \\ 
VGG net + Seq-NMS (avg) & 48.7 \\ 
VGG net + Seq-NMS (best) & 48.2 \\ \hline 
\textbf{CUVideo - T-CNN \cite{kang2016t}} & \textbf{67.8} \\ \hline
\end{tabular}
\end{table}

\section{Related Work}

Many previous works in video object detection framed as multiple object tracking. A popular subclass of these techniques were models that did "tracking-by-detection", whereby a detection algorithm is applied on each video frame and the detections are associated across frames to form trajectories for each object. Previous detection methods were usually based on motion \cite{zivkovic2004improved} or object appearance \cite{felzenszwalb2010object}. With regards to the association step, a classic method involved using Kalman filters to predict tracks and the Hungarian method \cite{perera2006multi, zhang2013understanding} to associate detections between frames. Particle filter techniques \cite{yang2005fast, breitenstein2009robust} further improved on Kalman filters by being able to handle multiple hypotheses. Other classes of methods tried to compute all of the object trajectories at once using linear programming \cite{jiang2007linear, berclaz2011multiple}. While these methods are able to find a global optimum with high probability, they assume that the number of objects to be tracked is known a priori. On the other hand, dynamic programming \cite{wolf1989finding, bercla2006robust} can also be used to find trajectories one by one in a greedy fashion. Our proposed model is similar in that it takes detections from a state-of-the-art single image object detection method \cite{ren15fasterrcnn} and subsequently associates tracks over time by finding the highest scoring path by also using dynamic programming. 

\section{Conclusion}
By using the strong baseline of Faster R-CNN and leveraging additional temporal information, we were one of the top performers in the ImageNet Object Detection from Video challenge. We would like to continue pursuing improvements to our submission, including training on the entire VID dataset, experimenting with neural network suppression, and performing a deeper analysis on our model designed to elucidate its weaknesses. 

\section*{Acknowledgments}
The six Tesla K40 GPUs used for this research were donated by the NVIDIA Corporation.

\small{
\bibliographystyle{plain}
\bibliography{mybib}

\begin{thebibliography}{10}

\bibitem{bercla2006robust}
J{\'e}r{\^o}me Bercla, Francois Fleuret, and Pascal Fua.
\newblock Robust people tracking with global trajectory optimization.
\newblock In {\em Computer Vision and Pattern Recognition, 2006 IEEE Computer
  Society Conference on}, volume~1, pages 744--750. IEEE, 2006.

\bibitem{berclaz2011multiple}
Jerome Berclaz, Francois Fleuret, Engin T{\"u}retken, and Pascal Fua.
\newblock Multiple object tracking using k-shortest paths optimization.
\newblock {\em Pattern Analysis and Machine Intelligence, IEEE Transactions
  on}, 33(9):1806--1819, 2011.

\bibitem{breitenstein2009robust}
Michael~D Breitenstein, Fabian Reichlin, Bastian Leibe, Esther Koller-Meier,
  and Luc Van~Gool.
\newblock Robust tracking-by-detection using a detector confidence particle
  filter.
\newblock In {\em Computer Vision, 2009 IEEE 12th International Conference on},
  pages 1515--1522. IEEE, 2009.

\bibitem{felzenszwalb2010object}
Pedro~F Felzenszwalb, Ross~B Girshick, David McAllester, and Deva Ramanan.
\newblock Object detection with discriminatively trained part-based models.
\newblock {\em Pattern Analysis and Machine Intelligence, IEEE Transactions
  on}, 32(9):1627--1645, 2010.

\bibitem{fastrcnn}
Ross Girshick.
\newblock Fast {R-CNN}.
\newblock In {\em Proceedings of the International Conference on Computer
  Vision ({ICCV})}, 2015.

\bibitem{jiang2007linear}
Hao Jiang, Sidney Fels, and James~J Little.
\newblock A linear programming approach for multiple object tracking.
\newblock In {\em Computer Vision and Pattern Recognition, 2007. CVPR'07. IEEE
  Conference on}, pages 1--8. IEEE, 2007.

\bibitem{kang2016t}
Kai Kang, Hongsheng Li, Junjie Yan, Xingyu Zeng, Bin Yang, Tong Xiao, Cong
  Zhang, Zhe Wang, Ruohui Wang, Xiaogang Wang, et~al.
\newblock T-cnn: Tubelets with convolutional neural networks for object
  detection from videos.
\newblock {\em arXiv preprint arXiv:1604.02532}, 2016.

\bibitem{perera2006multi}
AG~Amitha Perera, Chukka Srinivas, Anthony Hoogs, Glen Brooksby, and Wensheng
  Hu.
\newblock Multi-object tracking through simultaneous long occlusions and
  split-merge conditions.
\newblock In {\em Computer Vision and Pattern Recognition, 2006 IEEE Computer
  Society Conference on}, volume~1, pages 666--673. IEEE, 2006.

\bibitem{ren15fasterrcnn}
Shaoqing Ren, Kaiming He, Ross Girshick, and Jian Sun.
\newblock Faster {R-CNN}: Towards real-time object detection with region
  proposal networks.
\newblock In {\em Neural Information Processing Systems ({NIPS})}, 2015.

\bibitem{simonyan2014very}
Karen Simonyan and Andrew Zisserman.
\newblock Very deep convolutional networks for large-scale image recognition.
\newblock {\em arXiv preprint arXiv:1409.1556}, 2014.

\bibitem{selectivesearch}
J.R.R. Uijlings, K.E.A. van~de Sande, T.~Gevers, and A.W.M. Smeulders.
\newblock Selective search for object recognition.
\newblock {\em International Journal of Computer Vision}, 2013.

\bibitem{wolf1989finding}
Jack~K Wolf, Audrey~M Viterbi, and Glenn~S Dixon.
\newblock Finding the best set of k paths through a trellis with application to
  multitarget tracking.
\newblock {\em Aerospace and Electronic Systems, IEEE Transactions on},
  25(2):287--296, 1989.

\bibitem{yang2005fast}
Changjiang Yang, Ramani Duraiswami, and Larry Davis.
\newblock Fast multiple object tracking via a hierarchical particle filter.
\newblock In {\em Computer Vision, 2005. ICCV 2005. Tenth IEEE International
  Conference on}, volume~1, pages 212--219. IEEE, 2005.

\bibitem{viz_und}
Matthew~D. Zeiler and Rob Fergus.
\newblock Visualizing and understanding convolutional networks.
\newblock In {\em Computer Vision - {ECCV} 2014 - 13th European Conference,
  Zurich, Switzerland, September 6-12, 2014, Proceedings, Part {I}}, pages
  818--833, 2014.

\bibitem{zhang2013understanding}
Hongyi Zhang, Andreas Geiger, and Raquel Urtasun.
\newblock Understanding high-level semantics by modeling traffic patterns.
\newblock In {\em The IEEE International Conference on Computer Vision (ICCV)},
  December 2013.

\bibitem{edgeboxes}
C.~Lawrence Zitnick and Piotr Doll{\'a}r.
\newblock Edge boxes: Locating object proposals from edges.
\newblock In {\em ECCV}. European Conference on Computer Vision, September
  2014.

\bibitem{zivkovic2004improved}
Zoran Zivkovic.
\newblock Improved adaptive gaussian mixture model for background subtraction.
\newblock In {\em Pattern Recognition, 2004. ICPR 2004. Proceedings of the 17th
  International Conference on}, volume~2, pages 28--31. IEEE, 2004.

\end{thebibliography}
}

\end{document}